%% file: main.tex
\documentclass[journal,transmag]{IEEEtran}
\usepackage{cite}

\ifCLASSINFOpdf
\usepackage[pdftex]{graphicx}
\else
\fi
\usepackage{amsmath}
\usepackage{url}

\begin{document}
%
\title{NLP as a Lens for Causal Analysis and Perception Mining to Infer Mental Health on Social Media}


\author{\IEEEauthorblockN{Muskan Garg\IEEEauthorrefmark{1},
Chandni Saxena\IEEEauthorrefmark{2},
Usman Naseem\IEEEauthorrefmark{3}, Sohn Sunghwan\IEEEauthorrefmark{1}, and
Bonnie J Dorr\IEEEauthorrefmark{4}}
\IEEEauthorblockA{\IEEEauthorrefmark{1}Mayo Clinic, Rochester, MN, USA, garg.muskan@mayo.edu, sohn.sunghwan@mayo.edu}
\IEEEauthorblockA{\IEEEauthorrefmark{2}The Chinese University of Hong Kong, Hong Kong SAR, csaxena@cse.cuhk.edu.hk}
\IEEEauthorblockA{\IEEEauthorrefmark{3}The University of Sydney, NSW 2006, Australia, usman.naseem@sydney.edu.au}
\IEEEauthorblockA{\IEEEauthorrefmark{4}University of Florida, Fl, USA, bonniejdorr@ufl.edu}}

%



\IEEEtitleabstractindextext{%
\begin{abstract}
Interactions among humans on social media often convey intentions behind their actions, yielding a psychological language resource for Mental Health Analysis (MHA) of online users. The success of Computational Intelligence Techniques (CIT) for inferring mental illness from such social media resources points to NLP as a \textit{lens} for causal analysis and perception mining. However, we argue that more consequential and \textit{explainable} research is required for optimal impact on clinical psychology practice and personalized mental healthcare. To bridge this gap, we posit two significant dimensions: 1) Causal analysis to illustrate a cause-and-effect relationship in the user-generated text; 2) Perception mining to infer psychological perspectives of social effects on online users' intentions. Within the scope of Natural Language Processing (NLP), we further explore critical areas of inquiry associated with these two dimensions, specifically through recent advancements in discourse analysis. This position paper guides the community to explore solutions in this space and advance the state of practice in developing conversational agents for inferring mental health from social media. We advocate for a more explainable approach toward modeling computational psychology problems through the lens of language as we observe an increased number of research contributions in dataset and problem formulation for causal relation extraction and perception enhancements while inferring mental states. 
\end{abstract}

}

\maketitle

\IEEEdisplaynontitleabstractindextext

%
\IEEEpeerreviewmaketitle

\section{Introduction}

According to the World Health Organization (WHO) reports,\footnote{https://www.who.int/news/item/02-03-2022-covid-19-pandemic-triggers-25-increase-in-prevalence-of-anxiety-and-depression-worldwide} the prevalence of anxiety and depression is increased by 25\% in the first year of COVID-19 pandemic, yet many such cases have gone undetected. Traditionally, multiple in-person sessions with clinical psychologists are required to examine and infer a mental state, yet pandemic lockdowns affect the convenience of open sessions with mental health practitioners. 
Reports released in August 2021\footnote{https://www.theguardian.com/society/2021/aug/29/strain-on-mental-health-care-leaves-8m-people-without-help-say-nhs-leaders} indicate that \textit{1.6 million people} in England were on waiting lists for mental health care. 
An estimated \textit{8 million people} 
were unable to obtain assistance from a specialist,
as they were not considered \textit{sick enough} to qualify. This situation underscores the need for automation 
of mental health detection from social media data where people express themselves and their thoughts, beliefs/ emotions with ease. 
Motivated by~\cite{coppersmith2022digital}, there are continuously growing trends and patterns in this area of research which would benefit substantially from convenient
pathways and directions towards
explainability of AI models for mental health analysis on social media.\footnote{Explainability
refers
to the ability to determine reasons behind an algorithm's output
through generation of an explanation
for a particular decision, e.g., 
classification of
a social media post as an indicator of \textit{depression}.}

\begin{figure}
    \centering
    \includegraphics[width=0.5\textwidth]{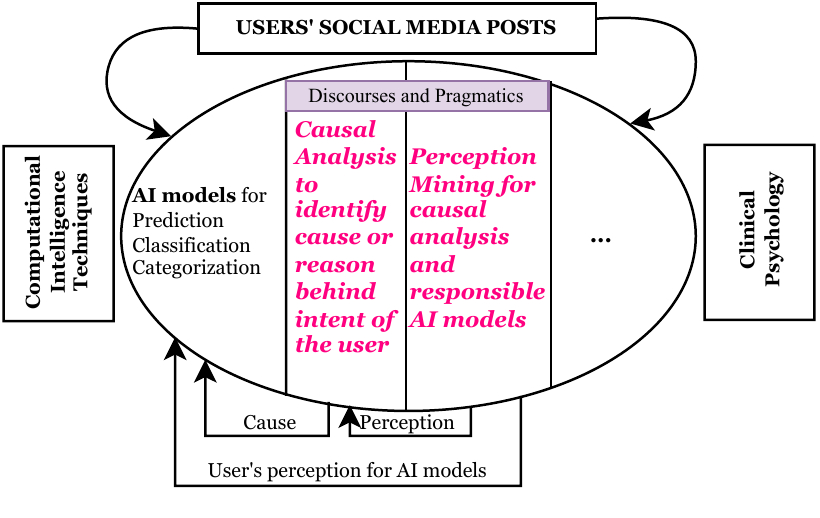} 
    \caption{To bridge a gap between computational intelligence techniques and clinical psychology, we pose \textit{causal analysis} and \textit{perception mining} by emphasizing on \textit{discourse and pragmatics} for social media analysis using NLP as a lens.}
    \label{fig:Fig0}
\end{figure}

Social media platforms are frequently relied upon as open fora for
honest disclosure. 
Social NLP researchers analyze social media posts to obtain useful insights for behavioral therapy. 
Facebook, 
a leading social media platform, uses \textit{artificial intelligence} and \textit{pattern recognition} to find users at risk.\footnote{https://www.cnet.com/tech/services-and-software/facebook-is-using-ai-to-scan-your-posts-for-suicidal-thoughts/}
The use of social media data for Mental Health Analysis (MHA) is bolstered by users' propensity for
self-disclosures, which often serve as a
therapeutic component of social well-being~\cite{jourard1959healthy}. 

\noindent \textbf{Motivation}: Motivated with the need to understand, process and generate human behavior for modeling real-time conversational AI agents, we examine the social NLP literature to process self-reported articles beyond syntactic and semantic analysis. To this end, we examine discourses and pragmatics in self-reported texts for inferring empathetic and perceptual understanding of mental states. Moreover, according to official 2019 report by Department of Veterans Affairs, 6,261 veterans die by suicide which is 7\% less than the previous year. However, as per \textit{national strategy for preventing veteran suicide: 2018-2028}, the Department of Veterans Affairs et. al., 2018 targets the national goal of reducing the number to 20\% by 2025. The adverse effect of isolation on veterans demands the need of personalized therapy through automated conversational AI agents.

\begin{figure*}
    \centering
    \includegraphics[width=0.80\textwidth]{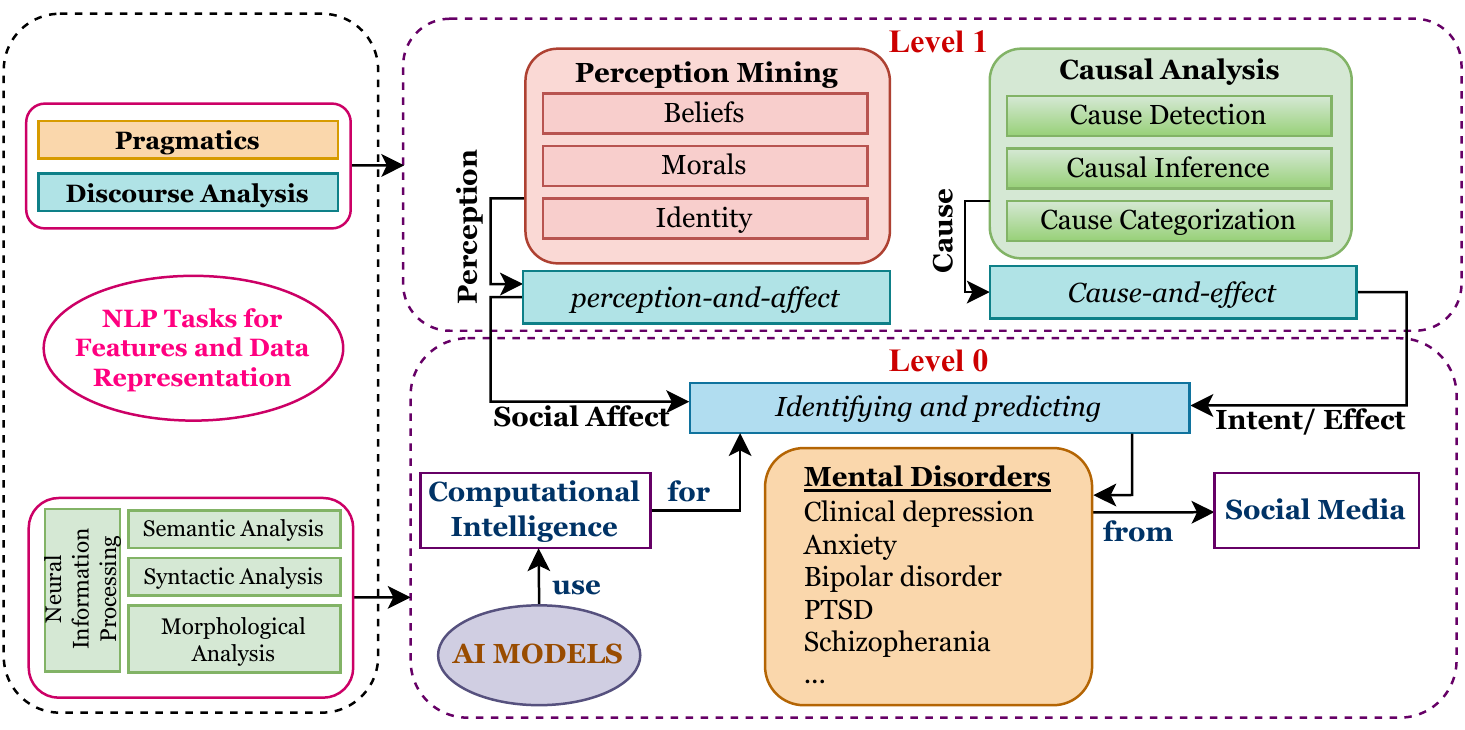}
    \caption{An overview of the current position on \textit{inferring mental health on social media}. We present the need of a Level~1 study (an in-depth study of pragmatics/discourse) for analyzing users' perception and causes behind their intent (mental illness) over existing Level~0 studies (intent prediction).}
    \label{fig:Fig2}
\end{figure*}


\noindent \textbf{Scope of this study}: Social media posts are \textit{cause-and-effect expositions} which may or may not contain the signs of reason behind the intent of a user.
Existing studies 
cover 
the significance of 
causal explanations inferred from
writing style~\cite{peterson1988pessimistic}. However, 
studies on the pragmatic use of languages and discourse analysis on social media are limited. To bridge this gap between Computational Intelligence Techniques (CIT) and clinical psychology, we pose \textit{causal analysis and perception mining} by emphasizing on \textit{discourse and pragmatics} in social media texts as shown in Figure~\ref{fig:Fig0}. 
We carry out collaborative discussions among the NLP research community and a senior clinical psychologist to maintain the integrity of this position paper. The scope of our work is limited
to the presentation of a new perspective. Our position paper moves this line of research in CIT one-step closer to the automation for real-time applications of clinical psychology. 
We identify a window of opportunity
for the NLP community to extract more nuances about human behavior, highlighting key challenges and future vision with an in-depth analysis of mental illness on social media.

\subsection{Current Position of Community}

Clinical psychologists conduct in-person sessions to understand human behavior for mental health analysis. This activity is simulated using Natural Language Understanding (NLU) over social media texts. However, 
research and publications in this area are still limited, hindering 
rapid community access to available resources 
to support such simulations.
Some interesting surveys and reviews are available on machine learning and deep learning for identifying and predicting the intent of affected users~\cite{rezapour2022machine, zhang2022natural}. The results have been convincing enough to embrace computational studies for MHA by projecting prominent key findings and their limitations to bridge a gap between two disciplines. 

Comprehensive studies of learning-based techniques have been applied to digital data~\cite{graham2019artificial}, medical record~\cite{eichstaedt2018facebook}, complex and large (original) reports~\cite{bernert2020artificial}, social media text~\cite{kim2021machine,d2020ai,calvo2017natural} and multimodal data~\cite{yazdavar2020multimodal}. Systematic studies and reviews towards this research field 
recapitulate the problem of demographic bias in data collection mechanism, managing consent of users for data, and theoretical underpinnings about human behavior in user-generated information~\cite{heckler2021machine}. 
Although such studies have made headway in human understanding of causes and perceptions associated with user-generated social media posts, automated identification of mental health related causes and perceptions is an area that has not yet come to fruition.

We 
classify this field of research into two levels 
of scientific study 
for
inferring mental health from social media language, as shown in Figure~\ref{fig:Fig2}:
\textit{Level~0} is the most prevalent position of the NLU community. Researchers in this realm build algorithms for mental health identification and
prediction, often relying on
handcrafted or automated features for building AI models~\cite{sawhney2021phase,ji2021mentalbert}. 
Contributions toward this endeavor have emerged through exclusive studies on computational intelligence with ethical research protocols, as it is mandatory to address ethical concerns due to the sensitive nature of datasets from which such algorithms develop~\cite{harrigian2021state,benton2017ethical,conway2016social}.

\textit{Level~1}, the perspective we rely upon in our position, assumes an in-depth analysis using \textit{perception-and-affect} and \textit{cause-and-effect} relationships. We view this as a new research direction
to build real-time explainable AI models. This line of research is aligned with the initial signs of
causal explanation analysis on Facebook data~\cite{son2018causal}, which has opened up new research directions to identify the reasons behind mental illness.








\subsection{Our Position}

As outlined above,
\textit{Level~0} studies 
contribute to a well-established line of research, but we posit that
\textit{Level~1} studies 
provide a basis for
an in-depth analysis of human behavior via \textit{causal analysis} and \textit{perception mining}. In light of these considerations, this position paper presents NLP as a lens 
through which to infer mental health on social media via two paradigmatic approaches: 
\begin{enumerate}[leftmargin=*]

    \item \textbf{Causal Analysis}: Users' social media posts may express their grief and reasons, thus providing
    background for their mental illness or justification for actions under consideration. The causal analysis is further classified as \textit{cause detection, causal inference and cause categorization}. Thus, \textit{Causal analysis} is a cross-sectional study 
that identifies reasons behind the intent of a user. 
    \item \textbf{Perception Mining}: The mental state of 
    users may be inferred from online postings of 
    perspectives expressed in a social media post. Identifying \textit{beliefs, morals, and identity} in the user's arguments lays a foundation for identifying and predicting mental states. Thus, \textit{Perception mining} deals with the way a user 
interprets sensory information 
that
affects the social attitude of a person. As such, perception mining acts as a backbone for causal analysis.
\end{enumerate}
We 
investigate the problem formulation, current position, and open research directions for causal analysis and perspective mining in Sections 2 and 
3, respectively.

\section{Causal Analysis}
A data-driven approach uses Pointwise Mutual Information (PMI) to find correlations between two \textit{verb phrases} acquired from data~\cite{chambers2008unsupervised}. Causal inference in place of correlation gives better and directional insights among different phrases~\cite{weber2020causal}. 
Causal analysis is an untapped area of research for MHA, probably due to its perceived difficulty. In this position paper, we 
adopt the view that
\textit{cause-and-effect relationships} have significance for MHA and, moreover, that exploration of cause-and-effect requires
discourse parsing beyond textual features.
Researchers in psychology have found that the human mind has a very complex mechanism for identifying and attributing the cause for their mental disturbance~\cite{khoo2002many}. Inferring cause-effect relations between intent of chronic problems such as depression, suicide risk and statements specifying reasons such as isolation, unemployment, has also been found to be an important part of usergenerated text comprehension, especially for narrative text.

\vspace{3pt}

\noindent \textbf{Terminology}: We introduce the \textit{intent} as an argument made by users on social media platforms while expressing their feelings, beliefs and circumstances. For this position paper, we further restrict the use of the term \textit{intent of a user} for arguments containing information about users' mental state only. Consider an example $A$ for a post written by a social media user as:

\begin{quote}
    A: I hate my job .. I cant stand living with my dad. I'm afraid to apply to any developer jobs or show my skills off to employers. I don't even own a car. I just feel like a failure.
\end{quote}

We then classify the \textbf{causal analysis} into three sub-tasks as indicated below, with the psychologist's input on concrete questions to constrain the nature of the reason behind intent of a user:
\begin{itemize}[leftmargin=*]
    \item \textbf{Cause Detection}: A classification technique to identify whether texts inferring users' mental health contains any \textit{reason or cause} behind \textit{user's intent}. Example $A$ shows that there exists at least one reason behind the poor mental state of a user (e.g., job, family issues, finances). Causal detection answers the question: "Does the text contain any indicator of the cause behind the mental condition, such as a job loss or a death in the family?".
    \item \textbf{Causal Inference}: An NLP task to obtain \textit{abstractive or extractive explanations} from a \textit{user's intent} after cause detection. Example $A$ reveals a causal inference as "\textit{hate my job, dont even have a car, hate my job, feels like failure}." Causal inference answers the question "Which parts of the text segments explains the reason behind mental illness?"
    \item \textbf{Cause Categorization}: Considering cause as a topic/concept, a topic-specific categorization of users' \textit{intent} using causal inference. A dominant cause in example $A$ is categorized as something related to \textit{Jobs and Career}. Causal categorization answers the question "Among given causal categories, which causal category does this text belong to?".
\end{itemize}

\begin{figure*}
    \centering
    \includegraphics[width=0.9\textwidth]{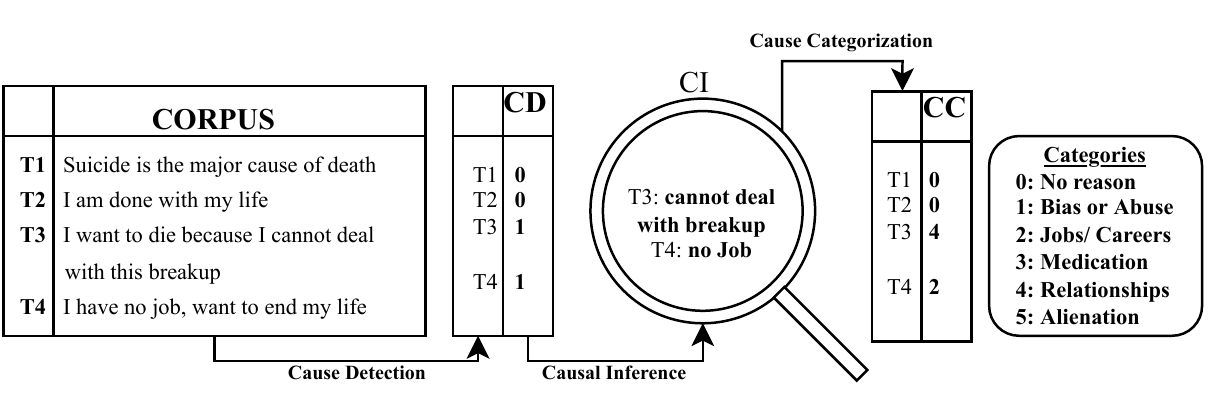}
    \caption{Working instance for causal analysis of user's intent in three steps: Cause Detection (CD), Causal Inference (CI) and Cause Categorization (CC). We give five categories as an example of cause categorization.}
    \label{fig:Fig3}
\end{figure*}
We briefly 
describe a working instance of a corpus, exploring 
three sub-tasks of causal analysis, as shown in Figure~\ref{fig:Fig3}. We 
use social NLP as a lens through which to conduct a
\textit{Level~1} study of \textit{intent}. 

\vspace{3pt}

\subsection{Psychological Theories}
We use NLP as a lens for the investigation of AI models developed for NLP tasks 
that categorize social media posts into their associated mental states. A \textit{neural Rhetorical Structure Theory (RST) parsing system} is publicly available to examine discourse relations and perceived persuasiveness of social media data~\cite{li2021neural}. Potential signs of cause behind
an imbalanced mindset are given in the posts such as insomnia, weight gain, or other indicators of worthlessness or excessive or inappropriate guilt. Underlying reasons may include:  
\textit{bias or abuse}~\cite{radell2021impact},
loss of \textit{jobs or career}~\cite{mandal2011job},
physical/emotional illness leading to, or induced by, \textit{medication} use~\cite{smith2015depression,tran2019depression},
\textit{relationship} dysfunction, e.g., marital issues~\cite{beach2002marital}, and
\textit{alienation}~\cite{edition2013diagnostic}. 
This list is not exhaustive, but it is a starting point for level~1 study of mental health analysis.

\subsection{Thinking beyond Social Features}
People with depression exhibit differences with respect to linguistic styles, such as the distribution of nouns, verbs and adverbs~\cite{gkotsis2016language}, resulting in the unconscious conceptualization of complex sentences. 
We advocate the use of behavioral features in the past such as \textit{first-person language}, \textit{present tense} and \textit{anger-based terms}~\cite{o2017linguistic}. Most of the existing language processing is associated with \textit{surface-level linguistic features} and \textit{semantic level aspects}.
for MHA on social media data.

Neural information processing uses automatic feature transformation in end-to-end models. 
Word embedding techniques such as Word2Vec, Glove and  FastText~\cite{cao2019latent} encode a token of the text in a dense vector representation.
More recently, 
 pre-trained language models such as BERT, GPT, Sentence BERT, and BART use attention mechanism to embed sentences and achieve state-of-the-art performance for cross-sectional studies\cite{lin2021survey}. 
Although neural information processing is suitable for some rapid assessments of mental state classification and categorization, 
data representation methods lack information
necessary to examine in-depth nuances of \textit{users' intent}. We briefly describe some 
possible solutions for the task of in-depth text analytics.

\subsection{Discourses for Causal Analysis}
The information extracted from textual features is in the form of the morphological, syntactic and semantic meaning of words from the \textit{intent of a user}. Causal analysis motivates the community to think beyond existing \textit{surface-level linguistic features} and \textit{semantic level aspects} yielding the need of \textit{level~1 studies} for mental health analysis.

\vspace{3pt}
\noindent\textbf{Knowledge Graph}: To “inject” mental disturbance through self-reported text into AI assistants such as Amazon Alexa, utilization of cross-domain knowledge of social interactions, emotions and linguistic variations of natural language is critical. Knowledge Graphs represents a network of real-world entities, namely, (i) objects as aspects of mental well-being, such as \textit{social aspect, vocational aspect, emotional aspect} (ii) events triggering mental disturbance, such as \textit{death, breakup, isolation} (iii) situations, such as \textit{human-user advocacy, domain knowledge, common-sense knowledge}. The illustration of relationships between them is visualized as a graph structure through a graph database. We map events from self-reported texts that indicate objects/ key aspects of mental disturbance through environmental situations suggesting the need of discourse analysis for mental healthcare. The complex nature of language processing tasks requires the construction of Knowledge-Graphs (KG) to capture text semantics~\cite{hogan2021knowledge}. KG's support the discovery of cause and effect relationships to reveal a reason behind suicidal intention~\cite{cao2020building}. We lay down a tuple to represent triplets as $<$\textit{event, object, relation}$>$ where \textit{event} is a reason that triggers mental disturbance and \textit{object} is the aspect of mental well-being thus affected through any given situational \textit{relation}. Such cause and effect relationships deduce discourse relations to examine reliability and hence, trustworthiness of decision making by AI models. 

\vspace{3pt}
\noindent\textbf{Discourse Relations}: There has been a recent surge in the use of discourse analysis, and its potential is demonstrated in a recent survey~\cite{drury2021survey}.
Discourse
analysis determines the connectivity among different text segments to map cause-and-effect relationships. 
Son et al.~\cite{son2018causal} conducts a recent experiment with Facebook data to extract causal explanations, which yields research insights for exploring discourse relations in the field of mental health analysis. More recent studies indicate that discourse relations support pragmatic inference worthy of future investigation in this domain as well \cite{son2021discourse}. 
Thus, Son et al.~\cite{son2018causal} introduce ground-breaking research
using discourse relations 
to detect mental health from Facebook data. The authors propose
an approach to \textit{cause detection} and \textit{causal inference} in their work and show promising results. However, the dataset is publicly unavailable and thus, address the limitations of dataset availability and the complexity of a problem.\\


\noindent Thus, discourse parsing and KG-based methods extract information from given social media post but without regard to its complexity. 
The longer the post, the more potential there is for introducing inconsistencies, further complicating the ability to understand user-generated language. A potential solution of complex self-reported text 
is \textit{sentence simplification}, 
rephrasing the sentence in a simplified form. 
~\cite{schwarzer2021improving}.
Many existing simplifications approaches 
rephrase the text without considering semantic information. However, to keep the essence of the cause-and-effect relationship, we argue that semantic dependency is required for this task. Semantic Dependency Information guided Sentence Simplification (SISS) is a neural sentence simplification system~\cite{lin2021neural}. 


\section{Perception Mining}
Clinical psychologists use their judgements to 
glean the psychological perception of a person via in-person offline sessions. Human judgements are more than common sense and regular language understanding. As a result, an in-depth analysis of self-reported social media posts is required to simulate human judgements about the psychological perspective of a user. 

In social media platforms, a historical timeline of users' posts reflects an overall attitude towards life. This attitude evolves from users' perceptions. A time-varying study, referred as \textit{longitudinal}, is used for identifying behavioral patterns to determine the extent to which a user is socially affected. 
Longitudinal studies enable the exploration of solutions to 
important research questions such as quantifying the social effect on a user, 
detection of suicidal ideation over a period of time, inferring changing patterns of mental health, and early risk prediction. Perspective mining supports solutions to these research questions. For this task, the given social media post
is analyzed using advanced stages of NLP: \textbf{pragmatics} and \textbf{discourse}. 



\vspace{3pt}

\begin{figure}
    \centering
    \includegraphics[width=0.5\textwidth]{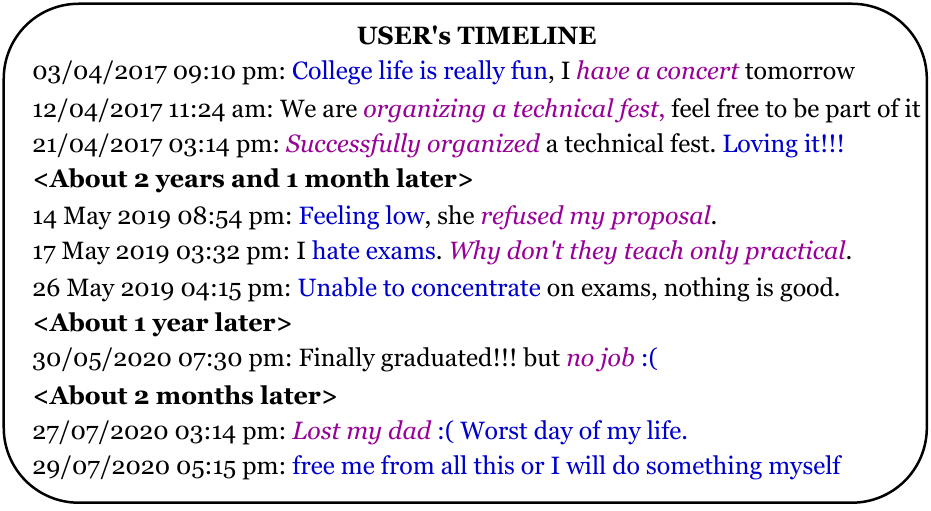}
    
    \caption{Users' historical timeline as an instance for perception mining. Overall attitude can be examined by tracking the user's historical timeline. User's perception of \textit{FREEDOM, ATTRACTION, ASSET} as evident from \textit{organizing a technical fest, refused proposal/ unable to concentrate, no job/ lost my dad}, respectively.}
    \label{fig:Fig4}
\end{figure}
\noindent \textbf{Working Instance}: Consider the timeline for a user as shown in Figure~\ref{fig:Fig4}. In given user's timeline, the purple and blue colored phrase indicates the \textit{cause} and \textit{intent} of a user, respectively. Our senior clinical psychologist suggests that the user's perception is \textit{FREEDOM, ATTRACTION, ASSET} as evident from \textit{organizing a technical fest, refused proposal/ unable to concentrate, no job/ lost my dad}, respectively. 
 From this, we see that perceptions uncover users' beliefs and morals which often underlie their intent as revealed through cross-sectional evaluation of causal analysis. Thus, we posit the need for perception mining to develop real-time explainable AI models such as conversational AI agents or AI chatbots for automatically handling mental health disorders.

 


\subsection{Psychological Theories} 
A person perceives through the five senses and sometimes through common sense as well to behave in a certain way. \textit{Self-perception theory} infers attitude and behavior via user's verbal and non-verbal actions~\cite{bem1972self}. We use NLP as a lens and consider written verbal communication. \textit{Structural balance theory} is a basic theory of cognitive consistency in social networks and examines the consistency in social behavior and user's attitude~\cite{hummon2003some}. There are many controversial topics, such as legal abortion, live-in relationships, early/ late marriages and joint/ nuclear family, on which people have different thoughts, beliefs and morals. An interesting theory on moral behavior sheds light on what people think about their identity~\cite{benabou2011identity}. With this background, we glean deep nuances and theoretical underpinnings of users' perceptions to understand them through AI models.

Taylor and Brown~\cite{taylor1988illusion} suggest that the social world and cognitive-processing mechanisms impose filters on incoming information and impact the psychological perspective of the user's well-being. A well-established study shows that a person's perception is a matter of pragmatics, depending largely on interpersonal relationships and their impact on quality of life~\cite{swann1984quest}. Correspondingly, a recent surge of pragmatics-inspired research on emotions has paved the way for new solutions to the problem of perception mining~\cite{wharton2022pragmatics,zhao2022find}.

\subsection{Investigating Personality in Text}

Although perception mining is closely related to psychological theories of personality, there is a significant difference between perception and personality detection. \textit{Personality} is a set of qualities/ characteristics which differentiates two persons and explains how they behave in society, but \textit{perception} is a way of organizing, identifying, and interpreting sensory information.\footnote{https://wikidiff.com/perception/personality} Thus, perception directly affects thoughts, actions, and behavior, it
is helpful 
to recognize situations and patterns. 
A user has
a change in perception due to 
perceptual aberrations, for instance, problems in perceiving cognitive information due to 
neurological disorders that
affect mental states. 
 

We further advocate the presence of recently introduced datasets and path-breaking models to examine perception of the author through language in social media. As evident from past studies, moralization in social networks provides useful insights about social 
phenomena
such as protest dynamics, message dissemination in network, and social distancing~\cite{mooijman2018moralization}.
Recent studies \cite{hoover2020moral} have
investigated $10$ categories of moral sentiment (care/harm, purity/degradation, etc.) in order to correlate stances in social media posts with both online and offline phenomena that include protest dynamics and social distancing.
Furthermore, perception mining 
supports extensive studies over discordant knowing~\cite{gollwitzer2022discordant} and  exploring social perceptions of health and diseases using social media data~\cite{fu2022casia}. 

\subsection{Pragmatics for Perception Mining}
Pragmatics deals with real-time situations sensibly and realistically in a way that is based on practicality rather than theoretical considerations. State-of-the-art NLP models acquire their knowledge of syntax, semantics and pragmatics from large amounts of text, on the order of billions of words, and store this knowledge in layers of artificial neural networks thereby addressing multiple long-standing problems in psychiatry~\cite{rezaii2022natural}. Existing well-equipped studies in pragmatic analysis of mental healthcare are empathetic conversations suggesting real-time application of online mental health support~\cite{saha2022towards,omitaomu2022empathic,ghosh2022persona,vu2022modeling}, and infusing commonsense knowledge~\cite{ghosh2022comma}. 





\section{Discussion}
The analysis above supports the critical need for 
automated analysis at a level that
supports mental health experts
understanding 
reasons and causes for mental-health states as expressed in social media posts. We posit that this is precisely where interpretable AI models are necessary for seeing beyond simple assignment of text snippets to mental-health categories, i.e., \textit{explainability}.

\subsection{Explainability}
Our perspective on level~1 studies encourage the NLP research community to find explanations behind the reflection of neuropsychiatric behavior in personal writings. Major challenges in advancing explainable AI for modeling user's behavior are (i) availability of limited dataset, (ii) quantitative and qualitative performance evaluation measures for user's perspective, and (iii) investigating discourse-specific explanations from long texts. We now depict a proposed representation as output for explainability with two examples for further illustration:

\begin{verbatim}
Text 1: …no point of living alone, my 
mother has no time for me!
\end{verbatim}

\begin{verbatim}
Text 2: Feeling low, she refused my proposal
Unable to concentrate on work.
\end{verbatim}
In Text 1, the mentions \textit{`no point of living’} and \textit{`mother has no time’} enable an inference that feeling neglected is a suicide risk indicator, and a perception of \textit{alienation} may lead to suicidal tendencies. The corresponding explainable representations are:\\

\begin{tabular}{lp{2.7in}}
    $\bullet$&\textit{causal\_relationship(neglect, suicide\_risk)}\\
     $\bullet$&\textit{perception\_mining(alienation, suicide\_risk)}
\end{tabular}\\

\begin{tabular}{lp{2.7in}}
    $\bullet$&\textit{causal\_relationship(rejection, depressed)}\\
     $\bullet$&\textit{perception\_mining(attraction, depressed)}
\end{tabular}\\

\begin{table*}[hbt!]
    \centering
    \small
    \hspace*{-0.4cm}
    \begin{tabular}{|p{0.4cm}|p{1.45cm}|p{4.5cm}|p{8.2cm}|}
        \hline
         \textbf{Out.} & \textbf{Dataset} &  \textbf{Task} & \textbf{Description}  \\
         \hline
         CA& CAMS & Causal Categorization \cite{garg2022cams} & Handling unstructured long texts to find reason behind intent. \\
         \hline
         CA& CAMS & Explainable NLP \cite{saxena2022explainable} & Explainable causal categorization of mental health. \\
        \hline
        PM & Twitter & Moral Foundation \cite{hoover2020moral} & Moral sentiment classification from Twitter data. \\ 
        \hline
        CA& Facebook & Causal Explanation \cite{son2018causal} & Causal explanation identification and extraction on social media. \\ 
        \hline
       PM & RHMD & Classification \cite{naseem2022rhmd} & Perception based health mention classification in Reddit posts. \\
       \hline
       PM & News & Empathy analysis \cite{omitaomu2022empathic} & Personality and belief driven empathetic conversation modeling. \\
       \hline
       PM & Twitter & Personality analysis \cite{giorgi2022regional} & Language-based personality assessment of regional users. \\
       \hline
       PM & Twitter & Beliefs \cite{vu2022modeling} & Modeling latent dimensions of human beliefs. \\
       \hline
       CA& CEASE & Causal Recognition \cite{ghosh2022cares} & Cause identification and extraction for emotions in suicide notes. \\
       \hline
       CA& Docs & ECPE \cite{chen2022learning} & Emotion-cause pair extraction (ECPE) from text documents. \\
       \hline
       PM & CEASE & Personality analysis \cite{ghosh2022persona} & Personality subtyping from suicide notes.\\
       \hline
       PM & MotiVAte & Dialogue system \cite{saha2022towards} & Empathetic response generation in online mental health support. \\
       \hline
       PM & Curated & Classify perception \cite{welch2022understanding} & Interpersonal conflict types for classifying perception. \\
       \hline
    \end{tabular}
    \caption{A list of initial tasks/entities is a resourceful compilation of references for causal analysis and perception mining of emotions, sentiments and thus, mental health. Here, \textit{Out} means ``Outcome'', \textit{CA} refers to ``Causal Analysis'', \textit{PM} refers to ``Perception Mining''.}
    \label{tab:my_label}
\end{table*}

In Text 2, the mentions \textit{'feeling low’} and \textit{‘concentration problem’} enable an inference that the author is depressed due to rejection, and a perception of \textit{attraction} leading to depression. The corresponding explainable representations are:\\

We posit that \textit{cause detection} must first be applied to determine whether there exists any cause, with binary output [0: does not exists, 1: does exists] for the author’s intent expressed in bold text. We further posit that \textit{causal inference} must then be applied to extract cause as an explanation in the author's phrases shown in italicized text. A final step categorizes the text into appropriate causes. We work with in-depth analysis of users’ perception to support causal inference and categorization. Other than these \textit{must-haves}, we leave a \textit{good-to-have} aspect to find correlations/ patterns among \textit{causal analysis} and \textit{perception mining}.

\subsection{Available Resources and Future Scope}

We enlist initial tasks/ entities as a resourceful compilation of references for causal analysis and perception mining in Table~\ref{tab:my_label}. We observe (i) publicly available datasets such as CEASE,\footnote{\footnotesize https://www.iitp.ac.in/\url{~}ai-nlp-ml/resources.html} CAMS,\footnote{https://github.com/drmuskangarg/CAMS} RHMD\footnote{https://github.com/usmaann/RHMD-Health-Mention-Dataset}, 
empathetic conversations\footnote{https://github.com/wwbp/empathic\_reactions}, 
(ii) dataset available on request such as MotiVAte \cite{saha2022towards}. Datasets curated in the past can be expanded with additional annotation and datasets for perception mining such as adding morals, values and beliefs. On the other hand, we come across three different problems formulated for causal analysis in the past as explained in \emph{Section 2.1}. We suggest a thought of problem formulation and data annotations for extending existing datasets to displace causal analysis on the top of perception mining and thus, reduce gap between the two.

Enriched with elements of (i) commonsense knowledge, (ii) domain-specific knowledge and (iii) other semantic enhancements for developing context-aware AI models for identifying, categorizing and predicting mental disorders, we more towards real-time applications such as developing conversational AI models through empathetic and personality analysis. We witness this low-level analysis through empathetic response generation~\cite{saha2022towards}, moral foundations~\cite{hoover2020moral}, semantic health mentions~\cite{naseem2022rhmd}, personality analysis~\cite{ghosh2022persona,omitaomu2022empathic}, human beliefs~\cite{vu2022modeling,omitaomu2022empathic} and cause-and-effect relationship in a given text~\cite{garg2022cams,ghosh2022cares,son2018causal,chen2022learning}. However, this high-level analysis misses key components to develop responsible AI models such as explainability, fairness, transparency, and accountability to deploy real-time applications in mental healthcare.

\section{Conclusion}
We posit \textit{causal analysis} and \textit{perception mining} for MHA on social media through the lens of NLP. The concept of causal analysis is described in three different stages: \textit{cause detection, extracting inference,} and \textit{cause categorization}. We examine existing textual features and introduce the need to exploit\textit{ discourse relations} and \textit{Knowledge Graphs (KG)} for causal analysis. Perception mining is an explainable feature for both AI models and causal analysis. The contribution of this work derives from the potential for tackling different use cases at a deeper, interpretable level than that of most existing approaches while addressing the ethical considerations required for developing real-time systems. We endeavor to disseminate this position widely in the research community and urge researchers to develop richer, explainable models for inferring mental illness on social media.


\section*{Ethical Considerations}
Although many anticipated research benefits are associated with our position above, the ethical implications of using NLP on social media text reveal a wide range of issues and concerns~\cite{laacke2021artificial}. Convey et al.~\cite{conway2014ethical} introduce a taxonomy of ethical principles on using Twitter in public health research. These ethical principles are applied on many social media platforms. In this section, we briefly highlight
ethical considerations for this line of research by examining different stakeholders involved and focusing on some important ethical principles, including \textbf{Privacy, Responsibility, Transparency} and \textbf{Fairness}.


\begin{itemize}[]
    \item \textbf{Privacy:} The research community experiences ethical challenges in ensuring data privacy on social media~\cite{conway2014ethical,bender2017ethics}. We adopt the guidelines of Benton et al. \cite{benton2017ethical}, which extend ethical protocols to guide NLP research from a healthcare perspective by framing privacy concerns for using social media data. Specifically, it becomes a privacy risk when personal attributes such as \textit{identity} of a person are revealed using publicly available data. For example, protecting users' (farmers/veterans) data privacy is essential, which is connected to their autonomy, personal identity, and well-being~\cite{reddy2020governance}.

    \item \textbf{Responsibility:}  
    Responsibility implies honesty and accountability in the application of CIT in mental health.
For example, it is the responsibility of healthcare professionals to ensure that CIT-based mental health applications provide benefits to users/patients, and it is the researcher's responsibility to design AI models to ensure ``traceability'' of decision-making processes. According to WHO\footnote{https://www.who.int/publications/i/item/9789240029200} guidance on \textit{Ethics and Governance of Artificial Intelligence for Health}, reliance on AI technologies in clinical care requires collective responsibility, accountability and liability among numerous stakeholders. Thus, a mindful practice has immense scope for the trustworthy and efficient exposition in mental health and improves clinical outcomes using such applications \cite{chancellor2019human,ismail2021ai,laacke2021artificial}. 

    \item \textbf{Transparency:}  The regulations of data transparency provided by a guidance note from the \textit{United Nations Development Group} address data collection challenges with due diligence.\footnote{https://unsdg.un.org/sites/default/files
/UNDG\_BigData\_final\_web.pdf} In this case, policymakers are concerned about developing a transparent data collection process that may ensure the confidentiality of users’ data. NLP researchers are responsible for transparency about computational research with sensitive data accessed during model design and deployment. 

    \item \textbf{Fairness:} Researchers are responsible for ensuring that the collected data are unbiased, balanced, and sufficient.
They are also accountable for better outcomes of NLP research for values like justice and equity. The development of fair AI technologies in mental healthcare supports unbiased clinical decision-making. Moreover,          \textit{interpretation}~\cite{song2018feature,aguilera2021detecting} and \textit{explanation}~\cite{jha2020explainable,uban2021explainability} are possible means for detecting bias so that it may be addressed. Furthermore, healthcare practitioners and researchers must collectively ensure effective evaluation mechanisms of AI technologies for mental health in support of trustworthy and fair decision-making.
\end{itemize}

\noindent In future, we encourage practical deployment of explainable and responsible AI models which adhere to ethical considerations.
\section*{Limitations}
The scope of our work is limited to abstract study and theoretical perspective of causal analysis and perception mining only. We acknowledge the absence of implementation/ empirical studies in this position paper and plan it for future work. 
Although there are no direct conclusions on the responsibility of AI from this perspective, 
we give cues about explainable AI in this work~\cite{beckers2022causal} and plan to carry out integrated studies with discourses in the near future. We limit our investigation to mining language in social media and avoid its extension to an in-depth study of clinical symptoms and diagnoses.


\input{output.bbl}

\end{document}